# Controlling Sparsity of Neural Network by Constraining Synaptic Weight on Unit $L_p$-Sphere

Weipeng Li, Xiaogang Yang, Chuanxiang Li, Ruitao Lu, Xueli Xie

Sparse deep neural networks have shown their advantages over dense models with fewer parameters and higher computational efficiency. Here we demonstrate constraining the synaptic weights on unit $L_p$-sphere enables the flexibly control of the sparsity with $p$ and improves the generalization ability of neural networks. Firstly, to optimize the synaptic weights constrained on unit $L_p$-sphere, the parameter optimization algorithm, $L_p$-spherical gradient descent ($L_p$SGD) is derived from the augmented Empirical Risk Minimization condition, which is theoretically proved to be convergent. To understand the mechanism of how $p$ affects Hoyer's sparsity, the expectation of Hoyer's sparsity under the hypothesis of gamma distribution is given and the predictions are verified at various $p$ under different conditions. In addition, the "semi-pruning" and threshold adaptation are designed for topology evolution to effectively screen out important connections and lead the neural networks converge from the initial sparsity to the expected sparsity. Our approach is validated by experiments on benchmark datasets covering a wide range of domains. And the theoretical analysis pave the way to future works on training sparse neural networks with constrained optimization.

Deep Neural Networks are among the hottest machine learning methods nowadays. They have led to major breakthroughs in various domains, such as computer vision[1], speech recognition, molecular biology, automatic medical diagnosis[2], and various predictions[3]. neural networks were originally designed to simulate the structure and function of the human brain. However, the dense connections used by mainstream architecture of neural networks is computationally intensive, easy to overfit, weak in generalization capability, and obviously inconsistent with the sparse connection mode of the human brain[4,5].

Inspired by the sparse topology of biological neural networks, sparse neural networks are presented and now well demonstrated on a variety of problems[6,7]. Many works have shown inference time speedups using sparsity for Recurrent Neural Networks[8] (RNNs), Convolutional Neural Networks[9] (CNNs), and Multi-Layer Perceptrons[10] (MLPs). It was concluded that pruning weights based on magnitude was a simple and powerful sparsification method in Ref. [11]. However, since the pruned network was lack of further training, the accuracy of the network is usually significantly reduced. To solve this problem, the idea of retraining the previously pruned network to increase accuracy was proposed in Ref. [12], but abundant additional training was required. To reduce the training epochs, pruning weights during the training of the network was proposed in Ref. [13], where connections were slowly removed over the course of a single round of training. Compared with the pure pruning, dynamic pruning and growing during training can significantly improve the accuracy and speedup the convergence of the sparse neural network. Sparse Evolutionary Training (SET)[10] proposed a simpler scheme where weights were pruned according to the standard magnitude criterion and were added back randomly. Single-Shot Network Pruning (SNIP)[16] attempted to find an initial mask to prune a given network at initialization and then use the connection sensitivity to decide which connection to keep during training. Sensitivity Driven Regularization (SDR)[14] introduced a regularization term to gradually reduce the absolute value of weights with low-sensitivity, where a large amount of weights gradually converged to zero, and could be safely deleted from the network. Rigging the Lottery (RigL)[17] updated the topology of the network during training by using weight magnitudes and infrequent gradient calculations. Properties of the sparse training techniques are summarized in Table 1.

Topology evolution is a key of sparse network training, which can be broadly divided into two categories: fixed sparsity during the training such as SET, SNIP and RigL, and adaptive sparsity depends on the complexity of data and network structure such as SDR. The first category precisely controls the sparsity of a neural network to every single layer, but it ignores the effect of the data distribution on the effective connections. Without careful selection, the preset sparsity for each layer is difficult to balance the need for fewer connections and higher fitting capabilities. The second category spontaneously adjusts the sparsity of a network during its training process. It tends to find a topology of minimum connections while maintaining an acceptable accuracy, but it is generally difficult to control the sparsity quantitatively. We believe that compared to the completely fixed sparsity and completely adaptive sparsity, a compromise may be a better choice: finding an indicator to adjust the sparsity while retaining the effect of data distribution, to ensure the fewest connections required for expressive features. This solution is expected to obtain a controllable sparsity where the sparsity of each layer is self-regulated and the overall sparsity is specified, to achieve higher accuracy.

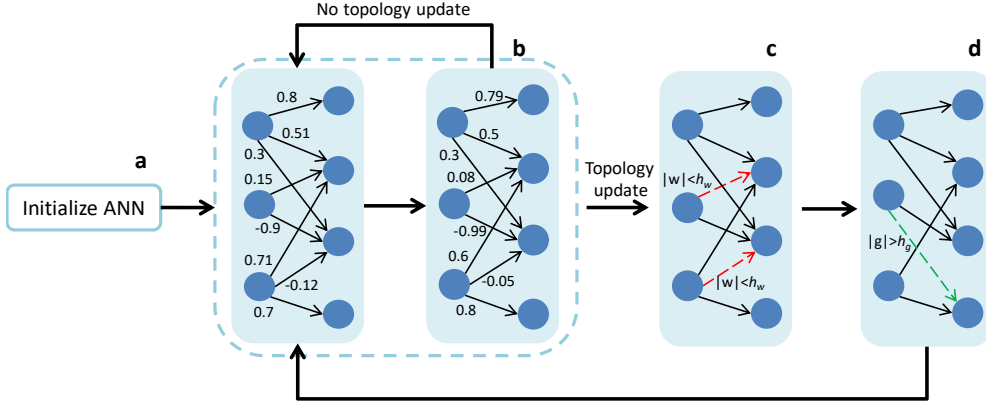

**Fig. 1 An illustration of the $L_p$SS procedure. a** Initialize neural network and $p$ to restrict the weights of each layer on $L_p$-sphere. **b** Train the weights with $L_p$SGD-m. **c** Drop the connections whose weight magnitude is lower than adaptive threshold. **d** Grow the connections whose gradient magnitude is larger than adaptive threshold. The process **b-d** continues for a finite number of training epochs.

| Table 1 Comparison of different sparse training methods | | | | |
|---|---|---|---|---|
| Method | Drop | Grow | Training | Space&FLOPs |
| SNIP | $\min(|\nabla_\theta L(\theta)|)$ | $\max(|\nabla_\theta L(\theta)|)$ | SGD | sparse |
| DeepR | stochastic | random | SGD | sparse |
| SET | $\min(|\theta|)$ | random | SGD | sparse |
| SDR | $|\theta|<h_\theta$ | $|\theta|\geq h_\theta$ | SGD | dense |
| SNFS | $\min(|\theta|)$ | momentum | SGD | dense |
| RigL | $\min(|\theta|)$ | $\max(|\nabla_\theta L(\theta)|)$ | $L_p$SGD | sparse |
| $L_p$SS(ours) | $|\theta|<h_w$ | $|\nabla_\theta L(\theta)|>h_g$ | SGD | sparse |
| Comparison of different sparse training techniques. Drop and Grow columns correspond to the strategies used during the mask update. | | | | |

$L_p$-norm has been successfully applied in feature selection[18], sparse representation[19], stabilization of Generative Adversarial Network[20], and sparse activation[21]. When the weights are constrained to the unit $L_p$-sphere, the searching space of weights will be restricted within a low-dimensional manifold, and the optimization algorithm mainly focuses on searching the orientations of weights[20]. We observe that under the constraint, Hoyer's sparsity of DNN increases with the decrease of $p$. Then we theoretically prove that under the gamma distribution, the expectation of Hoyer's sparsity of layers composed of independent neurons (e.g. convolutional layer) can be precisely predicted and controlled by $p$. Since Hoyer's sparsity is a soft approximation of standard sparsity, it is natural to think that training standard sparse DNNs by dropping the connections of weight close to zero.

In terms of parameter training, since the synaptic weights are constrained on the unit $L_p$-sphere while the classical gradient descent is used for unconstrained optimization, the restricted neural network cannot be trained by gradient descent. Therefore, a dedicated optimization method is needed. Based on the convexity of $L_p$-sphere, we propose the $L_p$-Spherical Gradient Descent ($L_p$SGD), an iterative form directly derived from the empirical risk minimization with the weight constrained on the $L_p$-sphere. To accelerate the training and improve the stability, $L_p$SGD with Momentum ($L_p$SGD-m) is proposed. it uses an additional momentum to accelerate $L_p$SGD along the relevant directions while reducing the oscillations in the irrelevant directions[23], and also provides an indirect update of weights to stabilize its direction and strengthen the control of Hoyer's sparsity.

In terms of topology evolution, since some weights are close to zero when trained by $L_p$SGD-m, their connections can be safely deleted from the network without reducing accuracy. Based on this idea, we propose the $L_p$-Spherical Sparse-learning ($L_p$SS), which uses adaptive dropping and growing to make the sparsity of the network gradually approach an expected value. Since RigL[17] presented an excellent topology evolution strategy for the sparsity fixed neural networks using the magnitude and the gradient of weights, we modified it to be an adaptive threshold strategy. Relying on our strategy, it avoids the deficiency of RigL that the sparsity of each layer needs to be carefully designed to balance the number of connections and fitting ability. On the other hand, the adaptive threshold avoids erroneous deletions of the connections whose weights are large occurring in dropping the minimum, thereby improving the stability of topology evolution. Finally, through the "semi-pruning" where drop more connections than growing, the neural network can screen out the decisive connections from the initial dense structure and improving the classification accuracy.

The rest of the paper is organized as follows. We first demonstrate the influence of $p$ on the distribution of weights constrained on the unit $L_p$-sphere and list the problems and corresponding solutions. Next, we introduce the theoretical details of $L_p$SGD and expectation of Hoyer's sparsity including their proofs and derivations. Then, we propose $L_p$SS to train sparse neural networks. Finally, numerous experiments are designed to verify the validity of our method in training the sparse neural networks on datasets covering a wide range of domains.

## Results
### Problem statement

We consider the neural network $f(x;\vartheta)$ with $L$ layers where the weight of each neuron is constrained on unit $L_p$-sphere. The corresponding model is given by:

$$\begin{aligned}\hat{y} &= f(x;\vartheta), \\ \text{s.t.} \quad &\| w_j^{(l)} \|_{p^{(l)}} = 1, l = 1, 2, \cdots, L, j = 1, 2, \cdots, m^{(l)},\end{aligned} \quad (1)$$

where $x$ is the input of neural network, $\hat{y}$ is the inference, $\vartheta = \{w, b\}$ is the set of weight and bias of neural network, $w_j^{(l)}$ is the synaptic weight of neuron $j$ in layer $l$, and $\| w \|_p = 1$ describes the vector $w$ is constrained on the unit $L_p$-sphere.

Fig. 2 illustrates the weight distribution of Eq.(1) with respect to $p$ under supervised learning. It can be observed that with the decrease of $p$ ($p>1$), the proportion of weights close to zero is increasing. Thus, we infer that the sparsity of neural networks can be controlled by $p$. However, since these weights are only close to zero but not exactly equal to zero, they cannot be properly evaluated by standard sparsity, which is defined as the proportion of zero components. We introduce an approximation of standard sparsity, Hoyer's sparsity[24] ($H_s$) to quantify the unbalance of weight distribution in a neuron. It is given by

$$H_s(w) = \frac{\sqrt{d} - \|w\|_1 / \|w\|_2}{\sqrt{d} - 1}, \quad (2)$$

where $w$ is a $d$-dimensional vector; $\|w\|_1$ and $\|w\|_2$ are the $L_1$-norm and $L_2$-norm of $w$. This function evaluates to unity if and only if $w$ contains only a single non-zero component, and takes a value of zero if and only if all components are equal in absolute value, interpolating smoothly between the two extremes.

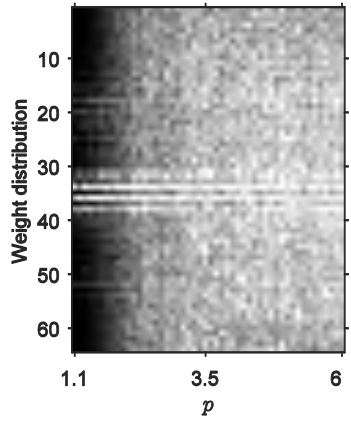

**Fig. 2 The weight distribution of a neuron in output layer with respect to *p*.** Each column is the synaptic weights of a neuron under certain *p*. The neural network is trained on MNIST dataset. It can be observed that with the decrease of *p* the proportion of weight close to zero is increasing.

After finding the approximate index of sparsity, we need to solve three problems for sparsity control:
(a) Since the weights in (1) are constrained and cannot be optimized by classical gradient descent, a dedicated optimizer is needed.
(b) To control Hoyer's sparsity by adjusting $p$, a theory is needed to explain the mechanism of how $p$ affects Hoyer's sparsity, and analyze the predictive power of this theory.
(c) To train a real sparse neural network, we need to explore how to transfer the control of Hoyer's sparsity to that of standard sparsity, and to propose a corresponding training algorithm.

For (a), $L_p$SGD is proposed to optimize the synaptic weights constrained on $L_p$-sphere and accelerate the convergence by using the convexity of $L_p$-sphere ($p>1$). It is theoretically proved to be convergent in form of partial differential equation by constructing a corresponding Lyapunov function, and be convergent in form of sequence while the gradient of empirical risk satisfies the Lipschitz continuity condition. To improve the stability of the training process and strengthen the control of Hoyer's sparsity, $L_p$SGD-m is proposed by combining $L_p$SGD with momentum.

For (b), the expectation of Hoyer's sparsity under the hypothesis that the input of each layer obeys the gamma distribution is derived. It is verified that the theory can precisely predict Hoyer's sparsity of weights for convolutional layers.

For (c), $L_p$-Spherical Sparse learning algorithm ($L_p$SS), a combination of $L_p$SGD-m and topology evolution with an adaptive threshold, is proposed to train the sparse neural network.

In our experiments, various sparse neural networks are trained on MNIST[25], Fashion-MNIST[26], CIFAR-10[27] and selected UCI machine learning datasets[28] without pre-trained model. Firstly, the optimization method $L_p$SGD and $L_p$SGD-m are tested on MNIST and Fashion-MNIST to train Hoyer's sparse neural networks with weights constrained on unit $L_p$-sphere. Then, in order to control Hoyer's sparsity of neural networks, the expectation of Hoyer's sparsity under gamma distribution is derived and compared with corresponding practical $H_s$ of each layer under various $p$ in neural networks trained on MNIST and Fashion-MNIST. Finally, $L_p$SS is introduced to train the real sparse neural networks, and is compared with some state-of-the-art methods in terms of sparsity-accuracy trade-off on various datasets.

**Training the weight constrained on unit $L_p$-sphere**

Gradient decent is one of most commonly used optimizer for unconstrained optimization. Since the supervised learning of (1) is a constrained problem, the classical gradient descent cannot be directly applied on it and there needs a dedicated optimizer. A constrained stochastic gradient descent (cSGD) for Riemannian optimization is proposed in Ref. [22]. It is a directly generalization of gradient descent by projecting the gradient on tangent space of Riemannian manifold. As a subset of Riemannian manifold, the unit $L_p$-sphere is certainly within the scope of cSGD. However, since $L_p$-sphere is convex when $p>1$, the property can be used to smoothen and accelerate the convergence during training, but cSGD is a general algorithm and does not use this property. Moreover, to ensure the versatility, cSGD calculates the normal vector of tangent space for all of weight vectors, which leads to a relatively high complexity.

Inspired by cSGD[22] and taking advantage of convexity of $L_p$-sphere when $p>1$, we propose the $L_p$-spherical gradient descent, an iterative form directly derived from the minimization condition of Empirical Risk with the weight constrained on $L_p$-sphere (see Method: $L_p$-spherical gradient descent). It is theoretically proved to be convergent in form of partial differential equation by constructing corresponding Lyapunov function, and be conditionally convergent in form of sequence (see Method: Convergence proof for $L_p$SGD). To improve the stability of training process, $L_p$SGD-m is proposed. It not only contains an additional momentum to accelerate the SGD along the relevant directions while reducing the oscillations in the irrelevant directions, but also provides an indirect update of weight to stabilize the direction of synaptic weights of each neuron thus strengthen the control of Hoyer's sparsity. (see Method: $L_p$SGD with Momentum).

Fig. 3 (on MNIST) and Fig. 4 (on Fashion-MNIST) show the training process of neural networks mentioned in **Table 2** using $L_p$SGD and $L_p$SGD-m. Although $L_p$SGD and $L_p$SGD-m have similar convergence process on test accuracy ($L_p$SGD and $L_p$SGD-m in Fig. 3 **a-d** and Fig. 4 **a-d**), the $L_p$SGD-m has obvious advantage in controlling Hoyer's sparsity with a relatively higher value and wider range on both of the MNIST and Fashion-MNIST. Especially when $p$ is small ($p=1.2$ in Fig. 3 **e** and Fig. 4 **e**), the $L_p$SGD-m obtains an obviously sparser neural network compared with $L_p$SGD.

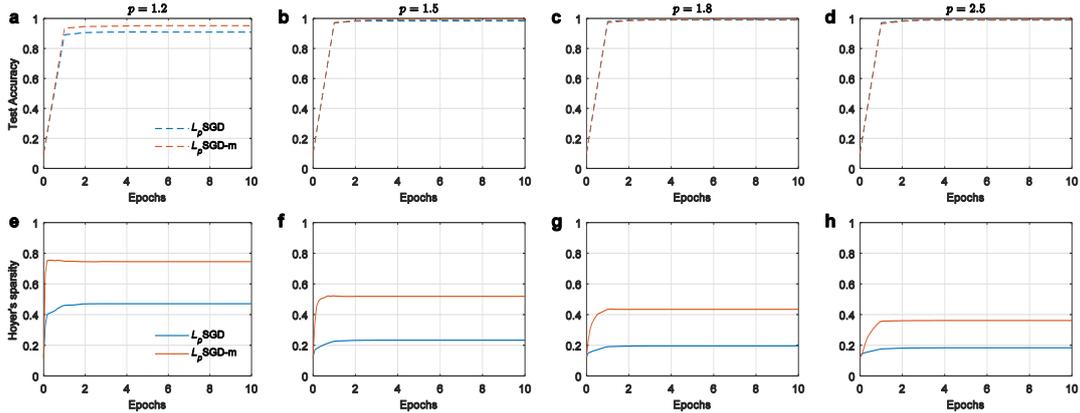

**Fig. 3 Training the neural network with $L_p$SGD and $L_p$SGD-m on MNIST. a-d** The training accuracy and test accuracy (y axis) over epochs (x axis). **e-h** Hoyer's sparsity (y axis) over epochs (x axis) corresponding to **a-d**. The neural networks are trained with learning rate starts at 0.02 which is scaled down by a factor of 0.3 every epoch. Each neural network is trained under different $p$: $p=1.2$ (**a, e**), $p=1.5$ (**b, f**), $p=1.8$ (**c, g**), and $p=2.5$ (**d, h**).

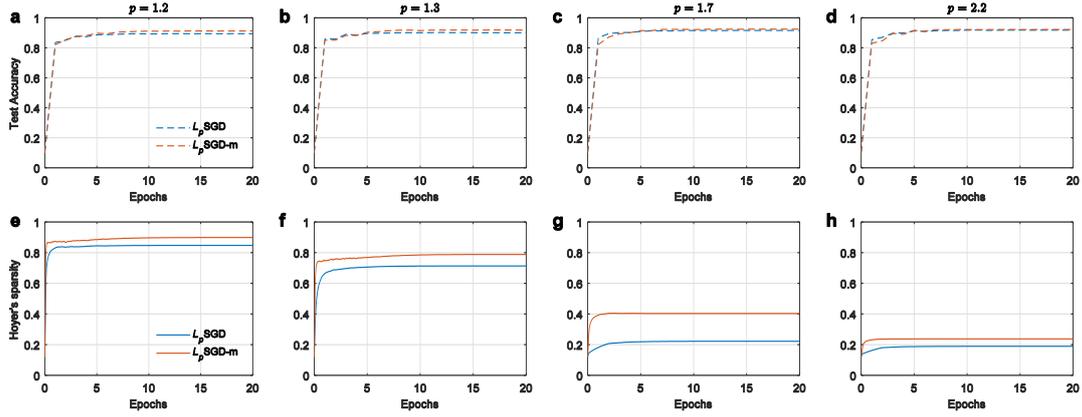

**Fig. 4 Training the neural network with $L_p$SGD and $L_p$SGD-m on Fashion-MNIST. a-d** The training accuracy and test accuracy (y axis) over epochs (x axis). **e-h** Hoyer's sparsity (y axis) over epochs (x axis) corresponding to **a-d**. The neural networks are trained with learning rate starts at 0.02 which is scaled down by a factor of 0.4 every 4 epochs. Each neural network is trained under different $p$: $p$=1.2 (**a**, **e**), $p$=1.3 (**b**, **f**), $p$=1.7 (**c**, **g**), and $p$=2.2 (**d**, **h**).

**Table 2 Structure of neural network for optimization comparison**

| Layer Type | MNIST | Fashion-MNIST |
|---|---|---|
| convolutional | 3×3 Conv. 8 neurons, ReLU | 5×5 Conv. stride 2, padding 2, 16 neurons, BN, ReLU |
| | 3×3 Conv. 12 neurons, ReLU | 3×3 Conv. padding 1, 32 neurons, BN, ReLU |
| | Max pooling stride 2 | Max pooling stride 2 |
| | 3×3 Conv. 16 neurons, ReLU | 3×3 Conv. padding 1, 64 neurons, BN, ReLU |
| | Max pooling stride 2 | |
| Fully connected | 400 Linear. 256 neurons, ReLU | 3136 Linear. 512 neurons, ReLU |
| | 256 Linear. 64 neurons, ReLU | 512 Linear. 64 neurons, ReLU |
| | 64 Linear. 10 neurons, output | 64 Linear. 10 neurons, output |

Conv. represents the convolutional filter, Linear. means the linear unit, BN is the batch normalization layer, all of the biases of convolutional layers before BN are disabled.

**Controlling Hoyer's sparsity**

As an approximation of standard sparsity, if Hoyer's sparsity can be controlled by adjusting $p$, it will be reasonable to believe that the standard sparsity can also be controlled by $p$ through simple topology evolution (such as drop the weights close to zero and grow some new connections). To understand the mechanism of how $p$ affect Hoyer's sparsity, we derive the expectation of $H_s$ with respect to dimension and $p$ under the hypothesis that the training data obeying gamma distribution, and then comparing the theoretical $H_s$ with practical $H_s$ under $p$ in different layers to justify the precision and the scope of theory. Since $p$ is not shared between layers, the $H_s$ of each layer is evaluated separately, and the practical $H_s$ are calculated as the average Hoyers' sparsity of weights over the neurons of neural networks trained on MNIST and Fashion-MNIST.

The details of neural networks are mentioned in **Table 2**. Both of the neural networks trained on MNIST and Fashion-MNIST have 6 layers, where the layers 1-3 are convolutional, and the layers 4-6 are full connected. To accelerate the convergence on Fashion-MNIST, each convolutional layer equips batch normalization.

Hoyer's sparsity ($H_s$) of weight with respect to $p$ at each layer is shown in Fig. 5 (trained on MNIST) and Fig. 6 (trained on Fashion-MNIST). Overall, the practical $H_s$ has an inverse relationship with $p$ in both of the convolutional layers and the fully connected layers, this matches our theoretical prediction. Compared with in fully connected layers, the practical $H_s$ of convolutional layers have wider range, and are more consistent with theoretical prediction. Moreover, we notice in fully connected layers, the practical $H_s$ are stable and are commonly higher than the predictions except for in $p<1.4$ on Fig. 5 **d** and $p<1.6$ on Fig. 6 **d**. This means Hoyer's sparsity of fully connected layer is intrinsic, and can only be affected instead of being completely controlled. It can be verified by the cross-correlations analysis of weights between the neurons of each layer in Fig. 7, where the cross-correlations of full connected layers except for the last layer, are obviously higher than that of convolutional layers under the same $p$. As a comparison, the neurons in convolutional layers (layer 1-3) are independent of each other, while the neurons in full connected layers (layer 4-6) are intrinsically correlated. The correlation in full connected clayers will lead to a compensation effect: when the weights of certain neuron changes, the weights of other positively or negatively correlated neurons will be adjusted to maintain the fitness of given output, leading to a stable sparsity. In summary, the theory can precisely predict Hoyer's sparsity of weight for convolutional layers, and can predict the tendency for full connected layers. Since the theory is more precise in the convolutional layers, it is expected to have better performace on CNNs.

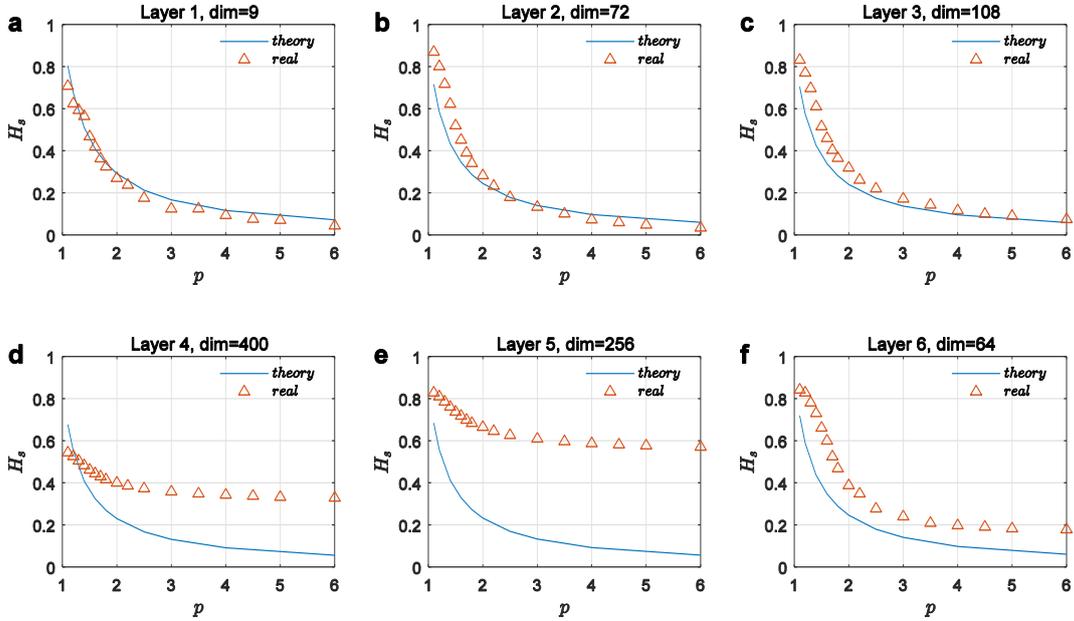

**Fig. 5 Hoyer's sparsity of weight with respect to *p* at each layer of neural network.** The blue curve (theory) is the theoretical value of $H_s$, the orange triangle is the practical sparsity (real) of weight trained on MNIST, and the "dim" means the dimension of the weights of each neuron. Layers 1-3 (**a-c**) are the convolutional layers, and layers 4-6 (**d-f**) are the full connected layers.

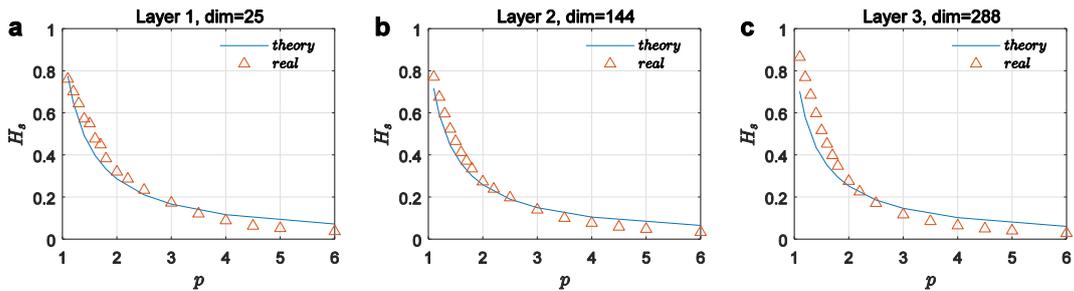

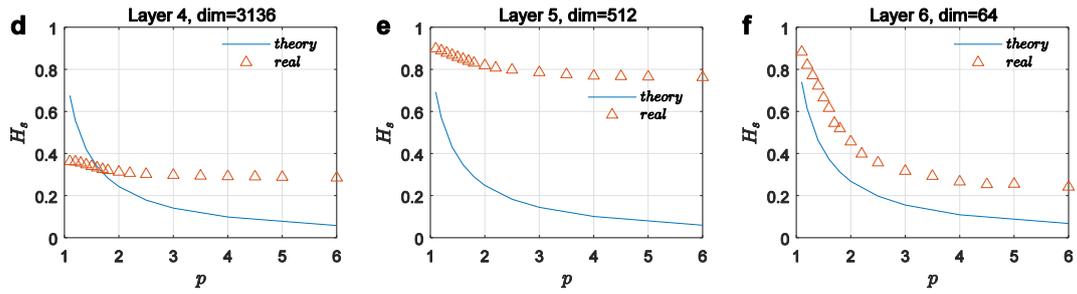

**Fig. 6 Hoyer's sparsity of weight with respect to *p* at each layer of neural network.** The blue curve (theory) is the theoretical value of $H_s$, the orange triangle is the practical sparsity (real) of weight trained on Fashion-MNIST, and the "dim" means the dimension of weights of each neuron. Layers 1-3 (**a-c**) are the convolutional layers, and layers 4-6 (**d-f**) are the full connected layers.

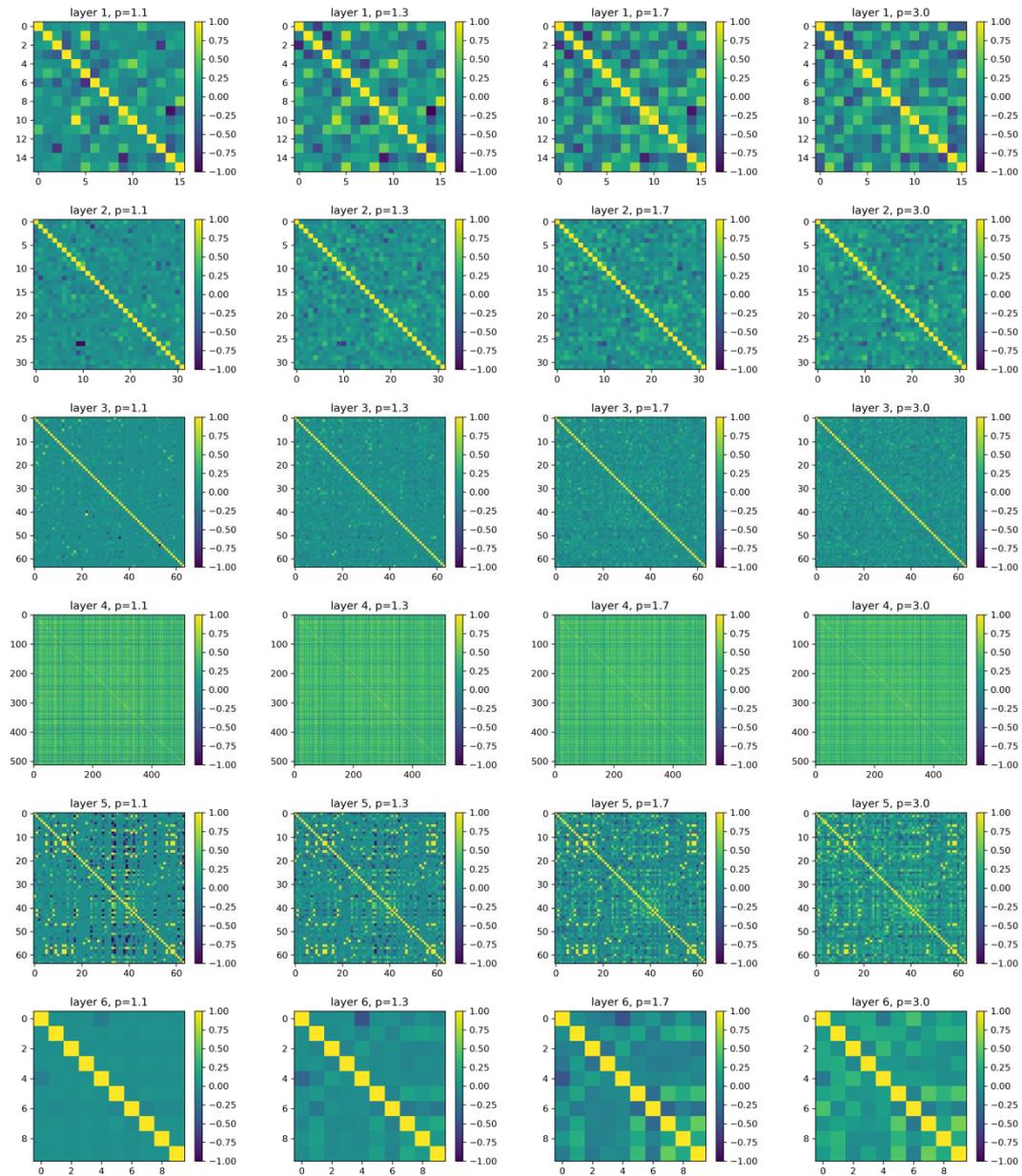

**Fig. 7 Part of cross-correlation matrix of weights between the neurons at each layer.** The neural network is trained on Fashion-MNIST. The figures form left to right are: *p*=1.1, *p* =1.3, *p* =1.7, *p* =3.0. And from top to bottom is the correlation matrixes from first layer to the last layer, where layers 1-3 are convolutional layers, and layers 4-6 are full connected layers.

**$L_p$-spherical sparse learning: From Hoyer's sparsity to standard sparsity**

Since the proportion of the weights close to zero is determined by $p$ under the $L_p$-spherical constraint, the sparsity of DNN can be controlled by adjusting $p$ and the threshold for eliminating the unimportant weights. With appropriate threshold adaptive adjustment strategy for dropping and growing, the sparsity of DNN can converge to the expected one. Based on this idea, we propose the $L_p$-Spherical Sparse learning algorithm ($L_p$SS) for sparse DNN, which is illustrated in Fig. 1. The main procedures of $L_p$SS are designed as follow: initialize the DNN and set $p$ for each layer, train the weights with $L_p$SGD-m, drop the weights below the threshold, grow the connections whose gradient higher than threshold, and update the thresholds with adaptive strategy. During the dropping and growing, we refer to RigL to use the magnitude of weight and the gradient of weight as indicators[17]. But we modify them into adaptive threshold determines instead of the maximum or minimum determines in RigL. The processing details are listed as follow.

(a) Initialize the DNN with small sparsity (0.1 or 0.2). For each layer, set $p$ to restrict the weights on unit $L_p$-sphere $\|w_j^{(l)}\|_{p^{(l)}} = 1$, where $l$ is the layer index, $j$ is the index of neuron.

(b) Train the neural network with $L_p$SGD-m algorithm introduced in Box 1.

(c) Every $\Delta t$ iterations, drop the connections whose weight magnitude is lower than the adaptive threshold: $\left\{w_{j,k}^{(l)}, \left|w_{j,k}^{(l)}\right| < \frac{\zeta_w}{n_{active}} \sum_{k \in W_{active}} \left|w_{j,k}^{(l)}\right|\right\}$, for each neuron $j$, where $\zeta_w$ is the relative threshold of weight, $W_{active}$ is the set of activated connections in this neuron, and $n_{active}$ is the size of $W_{active}$.

(d) At the same iteration of drop, grow the new connections whose gradient magnitude is highest in this neuron, $TopK\left\{\left|grad(w_{j,k}^{(l)})\right|, w_{j,k}^{(l)} \notin W_{active}\right\}$, where $K = \zeta_g n_{drop}$, $n_{drop}$ is the number of dropped connections in step (c) and $\zeta_g$ is the relative threshold for grow. Then the weights of newly activated connections are initialized to zero.

(e) Update $\zeta_w$ according to the update schedule (such as cosine annealing in (4)), and update $\zeta_g$ with the adaptive strategy

$$\zeta_g = \begin{cases} (1-\tau)s/s_{expect}, & s < s_{expect} \\ (1+\tau)s/s_{expect}, & s \geq s_{expect}, \end{cases} \quad (3)$$

where $\tau = 0.05$ is the gap between drop and grow, $s$ is the real sparsity, and $s_{expect}$ is the expected sparsity.

Finally, repeat (b)-(e) until the maximum iterations.

It should be noticed that, since different layers of neural network can be initialized with independent $p$, their sparsity can be controlled independently. This is especially important for the neural network whose input layer has few neurons that needs a larger $p$ to ensure the weights are expressive enough. In addition, since the effect of our constraint on sparsity is non-rigid, the sparsity cannot be completely controlled, thus $L_p$SS can only control the sparsity approximately instead of exactly equal to the specific value.

The performances of $L_p$SS on MNIST and Fashion-MNIST are shown in Fig. 8, where modified Wide Residual Network[29] (WRN) with 12 layers and a width multiplier of 1 (WRN-12-1) is trained on MNIST for 40 epochs, and the modified WRN-18-1 is trained on Fashion-MNIST for 50 epochs with a triangular learning rate. Each neural network is trained under different sparsity motivated by 4 groups of $p$. It can be observed that on both of MNIST and Fashion-MNIST, the neural network with nearly 0.5 sparsity generalize better than the dense baseline demonstrating the generalization ability of sparsity. With the increase of sparsity, test accuracy on both of MNIST and Fashion-MNIST slightly decreases.

We also evaluate the $L_p$SS on the CIFAR-10 image classification benchmark[27]. The basic learning rate used is presented in Fig. 9b, and the practiced learning rate is the basic one divided by batch size. It is designed to have several peaks and valleys to repeatedly reduce the sparsity by removing the connections which has little contribution to the accuracy during the peak, then growing a few necessary connections to reduce the loss during the valley. Finally, the sparsity will stabilize at the equilibrium point determined by $L_p$ constraint and data complexity.

The final accuracy of $L_p$SS under various sparsity levels is presented in Fig. 9a. The dense baseline obtains 94.1% test accuracy. The accuracy of SET, Static, RigL, SNIP, RigL[17] and $L_p$SS are obtained by 250 epochs of network training. With increased sparsity, we see a performance gap between the $L_p$SS and other solutions. Especially when sparsity is 0.9, the $L_p$SS achieves 93.1% test accuracy, which is significantly ahead of other methods. However, during the CIFAR-10 experiments we find a shortage of $L_p$SS: since the effect of $L_p$-sphere constraints to the sparsity is non-rigid, when sparsity increases to an upper bound that matches the complexity of training data and given $p$, the practical sparsity of the model will no longer converge to the expected final sparsity. That means if we want to increase the maximum sparsity reachable by $L_p$SS, the scale of the model itself should be increased. To explore the relationship between the model scale and the maximum sparsity, WRN-22-2, WRN-22-4, WRN-22-6 and WRN-22-8 are trained by $L_p$SS under the same $p=1.07$ (details are presented in [Implementation of WRN-22 training on CIFAR-10](#)), and the result is presented in Fig. 10. Generally, the maximum sparsity grows with the model scale, and the test accuracy increases with the logarithm of model scale. This implies the $L_p$SS would obtain sparser result when deployed to the models with larger scale. Moreover, since the networks are limited to be trained for 250 epochs, the WRN-22-8 has not achieved ideal convergence. We consider when the networks are fully convergent, the maximum sparsity and accuracy should have a linear relationship with the logarithmic model scale.

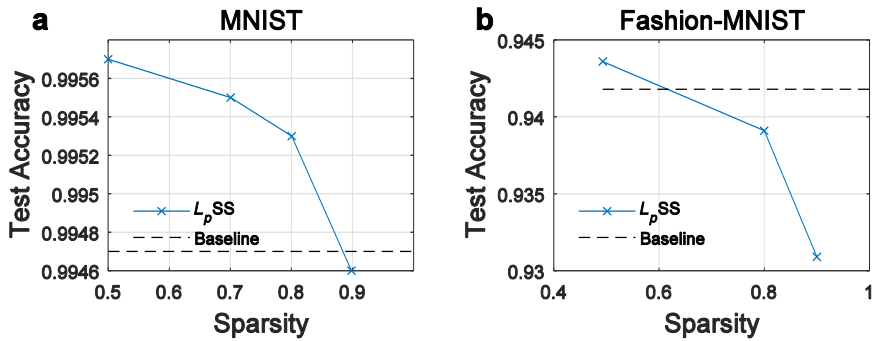

**Fig. 8 Test accuracy of neural network trained by $L_p$SS on MNIST and Fashion-MNIST under different sparsity.**
**a** Test accuracy with respect to sparsity on MINST. **b** Test accuracy with respect to sparsity on MINST. On both of **a**

and **b**, the blue curve is result of $L_p$SS, the black line (baseline) shows the accuracy of corresponding dense neural network.

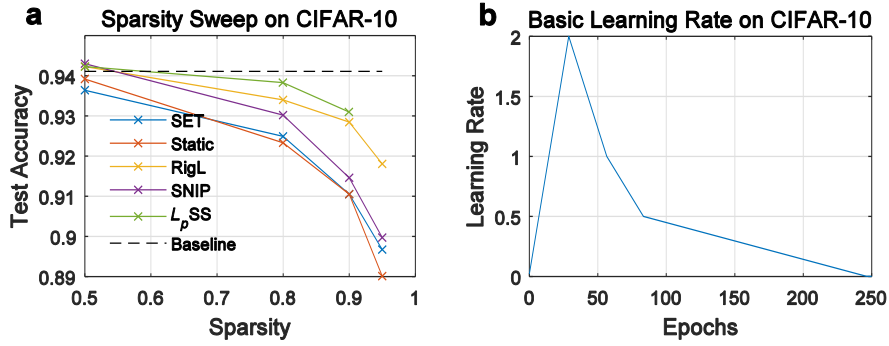

**Fig. 9 Training WideResNet-22-2 on CIFAR-10. a** Test accuracy of WRN-22-2 with respect to sparsity on CIFAR-10. **b** The basic learning rate used by $L_p$SS for training sparse WRN-22-2.

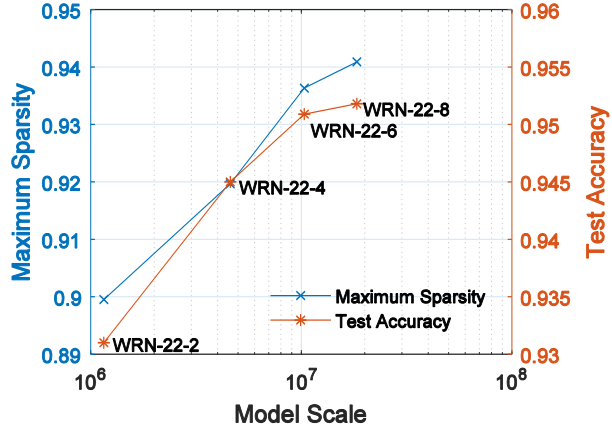

**Fig. 10 The effect of model scale on maximum sparsity that is approachable by $L_p$SS when *p*=1.07.** With the increase of model scale, the maximum sparsity and corresponding test accuracy increases.

**Performance on datasets in various domains**

We choose 11 benchmark datasets covering a wide range of domains, including social science, biology, physics, climate and compute vision, to evaluate $L_p$SS in training sparse DNNs. The details of the datasets are listed in Table 3. Since the data of selected UCI machine learning datasets are the vector of independent elements, the MLPs are suitable to deal with their classifications. While MNIST, Fashion-MNIST, CIFAR-10 and Tiny ImageNet are datasets of image, the CNNs are trained on them.

To obtain a slightly higher performance, cosine annealing strategy is introduced in update schedule for dropping connection in all of the algorithms in this experiment.

$$\zeta_w = \frac{\alpha}{2}\left(1+\cos\left(\frac{t\pi}{T_{end}}\right)\right), \tag{4}$$

where *α* is the initial relative threshold (or drop ratio) of weight, *t* is current iteration, and $T_{end}$ is the iteration stop connection update.

The test accuracy of each algorithm on datasets under different sparsity is mentioned in Table 4. In general, SET and Static performs better on MLPs, and this advantage is especially significant on UCI letter, while RigL, SNIP and $L_p$SS show advantage on CNNs, and this advantage is especially obvious on CIFAR-10 and Tiny ImageNet. $L_p$SS mainly outperforms other methods on CNNs: it achieves the highest accuracy in 7 of 9 most complex CNN models. Surprisingly, it also gets some impressive result on part of MLPs: it performs the highest accuracy in 7 of total 21 MLPs, and especially outstanding on DNA, Mushroom and Climate. In addition, $L_p$SS shows obvious advantage in 0.9 sparse DNNs (whether in CNNs or MLPs) by getting 6 highest accuracy on 11 datasets, which demonstrates the strong generalization ability of $L_p$-sphere constraint.

**Table 3 Datasets characteristics**

| Dataset | | Dataset properties | | | | |
|---|---|---|---|---|---|---|
| | | Domain | Data type | Features | Train samples | Test samples |
| UCI machine learning | Adult | Social | Binary&Float | 105 | 32,561 | 16,281 |
| | Connect-4 | Game | Binary | 126 | 57,557 | 10,000 |
| | DNA | Biology | Binary | 240 | 2,590 | 600 |
| | Mushroom | Biology | Binary | 126 | 6,124 | 2,000 |
| | Letter | Computer | Binary&Int | 16 | 15,000 | 5,000 |
| | Climate | Physics | Float | 18 | 400 | 140 |
| | Web | Computer | Binary | 27 | 1,153 | 200 |
| MNIST | | Computer | Grayscale | 784 | 60,000 | 10,000 |
| Fashion-MNIST | | Computer | Grayscale | 784 | 60,000 | 10,000 |
| CIFAR-10 | | Computer | RGB colors | 3072 | 50,000 | 10,000 |
| Tiny ImageNet | | Computer | RGB colors | 12288 | 100,000 | 10,000 |

The datasets in this paper cover a wide range of domains where DNNs have the potential to advance state-of-the-art, including social science, biology, physics, climate, website dataset and compute vision.

**Table 4 Summarization of test accuracy**

| dataset | | | s=0 | s=0.5 | | | | | s=0.8 | | | | | s=0.9 | | | | |
|---|---|---|---|---|---|---|---|---|---|---|---|---|---|---|---|---|---|---|
| | | $n^w$ | Baseline | SET | Static | RigL | SNIP | LpSS | SET | Static | RigL | SNIP | LpSS | SET | Static | RigL | SNIP | $L_p$SS |
| UCI machine learning | Adult | 323202 | 0.8605 | 0.8620 | **0.8626** | 0.8599 | 0.8588 | 0.8606 | 0.8618 | **0.8619** | 0.8614 | 0.8591 | 0.8452 | 0.7638 | 0.7638 | **0.8595** | 0.8510 | 0.7809 |
| | Connect-4 | 492291 | 0.8679 | **0.8675** | 0.8654 | 0.8649 | 0.8565 | 0.7898 | **0.8592** | 0.8570 | 0.8587 | 0.6558 | 0.7290 | **0.8489** | 0.8371 | 0.8203 | 0.6558 | 0.7172 |
| | DNA | 936195 | 0.9450 | 0.9483 | 0.9450 | **0.9583** | 0.9583 | 0.9533 | 0.5183 | 0.5183 | 0.8150 | 0.5183 | **0.9233** | 0.5183 | 0.5183 | 0.5183 | 0.5183 | 0.5183 |
| | Mushroom | 90562 | 1.0000 | 0.9890 | 1.0000 | 0.9985 | 1.0000 | 1.0000 | 0.5235 | 0.8915 | 0.9975 | 1.0000 | 0.9890 | 0.5235 | 0.5235 | 0.5235 | 0.5235 | **0.9755** |
| | Letter | 716378 | 0.9218 | **0.9710** | 0.9688 | 0.9688 | 0.9688 | 0.4546 | 0.9280 | **0.9600** | 0.9082 | 0.0368 | 0.3474 | 0.0368 | 0.0368 | 0.0368 | 0.0368 | **0.2328** |
| | Climate | 76738 | 0.9357 | 0.9500 | 0.9643 | 0.9500 | 0.9357 | **0.9714** | 0.9714 | 0.9214 | 0.9714 | 0.9214 | 0.9643 | 0.9214 | 0.9214 | 0.9214 | 0.9214 | 0.9214 |
| | Web | 212867 | 0.9050 | 0.9100 | **0.9350** | 0.9200 | 0.9150 | 0.8900 | 0.9250 | 0.9250 | 0.9250 | 0.9050 | 0.8400 | 0.4900 | 0.4900 | **0.8750** | 0.8700 | 0.8400 |
| MNIST | | 170298 | 0.9947 | 0.9955 | 0.9949 | **0.9959** | 0.9955 | 0.9957 | **0.9954** | 0.9944 | 0.9948 | 0.9941 | 0.9953 | **0.9947** | 0.9942 | 0.9943 | 0.9933 | 0.9946 |
| Fashion-MNIST | | 267514 | 0.9418 | 0.9432 | 0.9389 | 0.9420 | 0.9364 | **0.9436** | 0.9373 | 0.9357 | 0.9358 | 0.9281 | **0.9391** | 0.9256 | 0.9244 | **0.9316** | 0.9229 | 0.9309 |
| CIFAR-10 | | 1154074 | 0.9411 | 0.9364 | 0.9392 | 0.9423 | **0.9430** | 0.9423 | 0.9249 | 0.9233 | 0.9340 | 0.9302 | **0.9383** | 0.9105 | 0.9105 | 0.9285 | 0.9146 | **0.9310** |
| Tiny ImageNet | | 4644120 | 0.6138 | 0.5930 | 0.5924 | 0.6018 | 0.6049 | **0.6377** | 0.5686 | 0.5647 | 0.5815 | 0.5678 | **0.6295** | 0.5437 | 0.5419 | 0.5582 | 0.5264 | **0.6042** |

$n^w$ represent the number of weight, the bold number indicates the optimal accuracy of the group under same sparsity. The MLPs are trained on UCI machine learning datasets, while CNNs are trained on the MNIST, Fashion-MNIST, CIFAR-10 and Tiny ImageNet datasets.

Through Table 4, we find $L_p$SS seems to be sensitive to the model scale and dataset complexity: it performs well on larger models that over-fit the dataset, while performs less satisfactorily on models with insufficient fitting capabilities. We speculate the reason is that under the constraint, the degree of freedom of weight is reduced compared with the unconstrained model, leading to the requirement of additional parameters to achieve the same fitting ability. In addition, the reason of relatively lower performance on MLPs may be relate to the constraint of weight: according to Result: [Controlling Hoyer's sparsity](#), the $H_s$ of fully connected layer is intrinsic and can only be affected instead of being completely controlled under the constraint, when $H_s$ is small, the weights that obviously contribute to the accuracy might be dropped to achieve the expected sparsity. This is particularly significant when $p$ is large but the assigned final sparsity is small, the dropped weight will have larger magnitude and influence. We speculate the problem can be solved by optimizing the convolutional layers and full connected layers with $L_p$SGD and $L_p$-regularized SGD respectively.

## Discussion

In this paper, we study the training of sparse DNN, and introduce $L_p$SS, a semi-pruning method for training DNN from dense to sparse. We first demonstrate the influence of $p$ on the distribution of weights constrained on unit $L_p$-sphere and list the problems to be solved for training sparse DNN. We then solve these problems and propose the $L_p$SS which combines $L_p$SGD-m with adaptive threshold topology evolution. Through comparing with some state-of-the-art methods by training sparse DNNs on datasets in various domains, the effectiveness of $L_p$SS is demonstrated. It outperforms contrast methods in majority of CNNs and obtains impressive results on MLPs.

We model the supervised learning of DNN with $L_p$-spherical constrained weight into constrained Empirical Risk Minimization (ERM), and consequently derive the iterative solving algorithm $L_p$SGD according to the augmented ERM condition. The convexity of $L_p$-sphere when $p>1$ accelerates the convergence of constrained training with $L_p$SGD. It is theoretically proved to be convergent in form of partial differential equation by constructing corresponding Lyapunov function, and convergent in form of sequence when the gradient of empirical risk satisfies the Lipschitz continuity condition. To improve the stability of training process, $L_p$SGD-m is proposed, which not only contains an additional momentum to accelerate the SGD along the relevant directions while reducing the oscillations in the irrelevant directions, but also provides an indirect update of weight to stabilize the direction of weight of each neuron thus strengthen the control of Hoyer's sparsity.

Through the coordinate transformation to hypersphere, we deduced the theoretical expectation of Hoyer's sparsity with respect to $p$ under the hypothesis that the input satisfies the gamma distribution. By training the DNNs with various $p$ for each layer, we show the theory can precisely predict Hoyer's sparsity of weight for convolutional layers, and can predict the tendency for full connected layers. Moreover, we justify Hoyer's sparsity of fully connected layer is intrinsic, and can only be affected instead of being completely controlled, which is verified by the cross-correlations analysis of weights between the neurons of each layer.

Since the proportion of the weights close to zero is determined by $p$ when using $L_p$SGD-m, the sparsity of DNN depends on $p$ and the threshold eliminating the unimportant weights. In $L_p$SS, we design the threshold adaptive adjustment strategy for dropping and growing to evolve the sparsity of DNN to the expected one. It trains the sparse DNNs by repeating the procedures that train the weights with $L_p$SGD-m, drop the weights below the threshold, grow the connections whose gradient higher than threshold, and update the thresholds with adaptive strategy. Through the "semi-pruning" of grow less and drop more, $L_p$SS screens out the decisive connections from the initial dense structure and improves the accuracy.

We perform experiments on benchmark datasets covering a wide range of domains including social science, biology, physics, climate, and compute vision, to evaluate $L_p$SS in sparse DNN training. The experiment is a comprehensive test which trains CNNs and MLPs in different scales and sparsity with training data in various domains. It is a good test since it allows us to understand how model type and model scales affect the fitness and generalization ability of these training methods. And the experimental results largely match our prediction: $L_p$SS outperforms other methods on large scale CNNs, it achieves the highest accuracy in 6 of 9 large scale CNN models; it also shows obvious advantage in very sparse DNNs (with 0.9 sparsity), which demonstrates its strong generalization ability. Moreover, we find $L_p$SS is sensitive to the model scale and dataset complexity: it performs well on larger models that over-fit the dataset, while performs less satisfactorily on models with insufficient fitting capabilities. We speculate the reason is that under the constraint, the degree of freedom of weight is reduced compared with the unconstrained model, leading to the requirement of additional parameters to achieve the same fitting ability. On the other hand, the decline in fitting ability exchanges for the increase of generalization ability of $L_p$SS, which makes it more suitable for training the large scale sparse models overfitting the training data.

However, the $L_p$SS has relatively lower performances on MLPs compared to CNNs. We speculate this may be relate to the constraint of weight: since the neurons in full connected layers are correlated with each other under the constraint, the $H_s$ of that is intrinsic and can only be affected instead of being completely controlled, when $H_s$ is small, the weights that obviously contribute to the accuracy might be dropped to achieve the expected sparsity. This is particularly significant when $p$ is large while the assigned final sparsity is small, the dropped weight will have larger magnitude and influence. We believe the problem can be solved by respectively optimizing the convolutional layers and full connected layers with $L_p$SGD-m and $L_p$-regularized SGD. In addition, since $L_p$SGD-m has the ability of controlling the proportion of unimportant parameters while restricting them on unit $L_p$-sphere, we hope it can be applied to other energy restricted applications such as wireless sensor networks and generative adversarial networks.

## Methods
### $L_p$-spherical gradient descent

The neural network defined in (1) can be trained with Empirical Risk Minimization (ERM) under the constraint of unit $L_p$-norm of weight

$$\min_{\vartheta} R = \frac{1}{n} \sum_{i=1}^{n} \mathcal{L}(\hat{\boldsymbol{y}}_i, \boldsymbol{y}_i; \vartheta) \\ \text{s.t.} \quad \left\| \boldsymbol{w}_j^{(l)} \right\|_{p^{(l)}} = 1, l = 1, 2, \cdots, L, j = 1, 2, \cdots, m^{(l)}, \tag{5}$$

where $\mathcal{L}(\hat{\boldsymbol{y}}_i, \boldsymbol{y}_i)$ is the loss function of network about $i$-th training data. $\hat{\boldsymbol{y}}_i$ and $\boldsymbol{y}_i$ are the prediction and label respectively.

The corresponding augmented ERM is

$$\min_{\vartheta} R^* = R + \sum_{l=1}^{L} \sum_{j} \frac{\lambda_j^{(l)}}{p^{(l)}} (\| \boldsymbol{w}_j^{(l)} \|_{p^{(l)}}^{p^{(l)}} - 1), \tag{6}$$

where $\lambda_j^{(l)}$ is the Lagrange multiplier about $\boldsymbol{w}_j^{(l)}$.

Based on the equation that the gradient of $R^*$ with respect to $w_j^{(l)}$ is equal to zero under minimum Empirical Risk, the optimal weight is given by

$$\lambda_j^{(l)} \left(w_j^{(l)}\right)^{[p^{(l)}-1]} = -\nabla_{w_j^{(l)}} R\left(w_j^{(l)}\right), l = 1, 2, \cdots, L, \tag{7}$$

where $\nabla_w R$ is the gradient of empirical risk with respect to $w$. And

$$w^{[p]} = \text{sgn}(w) \circ |w|^p, \tag{8}$$

represent the Hadamard product of $p$-power of absolute value and corresponding sign for each element of $w$, where the components satisfies $w_j^{[p]} = \text{sgn}(w_j) |w_j|^p$.

To constraint $w$ on the unit $L_p$-sphere, $\lambda_j^{(l)}$ should be

$$\lambda = \|\nabla_w R(w)\|_q, \tag{9}$$

where $q = p/(p-1)$.

Set

$$\Delta(w) = \left[\frac{\nabla_w R(w)}{\lambda}\right]^{[q-1]}, \tag{10}$$

as the normalized gradient of $R$ with respect to $w$.

Since the solution of ERM satisfies equation (7), combined with (10), the optimal weight can be given by

$$w_j^{(l)} = -\Delta(w_j^{(l)}). \tag{11}$$

Set $\eta$ is the learning rate of weight, the expanded form of (11) is

$$w_j^{(l)} = (1-\eta)w_j^{(l)} - \eta\Delta(w_j^{(l)}). \tag{12}$$

Then the iteration of weight is given by

$$w_j^{(l,t+1)} = \frac{1}{\rho_j^{(l,t+1)}}\left[(1-\eta)w_j^{(l,t)} - \eta\Delta(w_j^{(l,t)})\right], \tag{13}$$

where $\rho_j^{(l,t+1)} = \|(1-\eta)w_j^{(l,t)} - \eta\Delta(w_j^{(l,t)})\|_{q^{(l)}}$ is the normalization factor of $w_j^{(l,t+1)}$. $w_j^{(l,t)}$ is the value of $w_j^{(l)}$ at iteration $t$, and $w_j^{(l,t+1)}$ is the value of that at iteration $t+1$.

The iteration of bias is a classical gradient descent

$$b_j^{(l,t+1)} = b_j^{(l,t)} - \eta \nabla_{b_j^{(l)}} R(b_j^{(l,t)}), \tag{14}$$

where $\nabla_{b_j^{(l)}} R(b_j^{(l,t)})$ is the gradient of empirical risk with respect to bias $b_j^{(l)}$ at iteration $t$.

It should be noted that the vanishing gradient which disturbs the classical neural network can be partially overcome by $L_p$SGD. The reason is that in iteration (13), the gradient $\nabla_{w_j^{(l,t)}} R(w_j^{(l,t)})$ is normalized by coefficient $\lambda_j^{(l,t)}$ which compelling the $p$-norm of $\Delta(w_j^{(l,t)})$ equals to 1, thus no matter how small the gradient is, it always contributes $\eta$ rate of updated weight during the training process.

Next we will prove the convergence of $L_p$SGD.

**Convergence proof for $L_p$SGD**

Since the iteration of bias is the classical gradient descent (GD), and the convergence of GD has already been proven[30], it will not be repeated in this paper. Here we focus on the convergence of weight. Firstly, we prove the convergence of partial differential equation (PDE) form of (13), which is given by

$$\dot{w} = \lim_{\eta \to 0} \frac{1}{\eta} \left( \frac{1}{\rho(\eta)} [(1-\eta)w - \eta \Delta(w)] - w \right)$$
$$= w \left( w^{[p-1]} \right)^T \Delta(w) - \Delta(w). \quad (15)$$

**Theorem 1.** Suppose $w^*$ is a local optimal solution of $R(w)$ on unit $L_p$-sphere (that means the connected set $\mathcal{O} \subseteq \{ w \mid \|w\|_p = 1 \}$ exists, where $R(w) > R(w^*), \forall w \in \mathcal{O} \setminus w^*$ ), then $w^*$ is an asymptotically stable equilibrium point of PDE (15).

*Proof.* According to the equation (11), the optimal solution of $R(w)$ on unit $L_p$-sphere is given by

$$w^* = -\Delta(w^*).$$

Constructing a Lyapunov function

$$\varepsilon(w) = R(w) - R(w^*).$$

It is easy to verify that this function satisfies the condition of Lyapunov function: $\varepsilon(w^*)=0$ and $\varepsilon(w) > 0, \forall w \in \mathcal{O} \setminus w^*$. Now if we can prove $\dot{\varepsilon}(w) < 0, \forall w \in \mathcal{O} \setminus w^*$, $w^*$ will be the asymptotically stable equilibrium point of PDE (15).

Since $\varepsilon(w)$ has been given, combined with $w$ in (15) the derivation of $\varepsilon(w)$ can be obtained

$$\dot{\varepsilon}(w) = \nabla_w R(w)^T \dot{w}$$
$$= \nabla_w R(w)^T w \left( w^{[p-1]} \right)^T \Delta(w) - \|\nabla_w R(w)\|_q.$$

Since $p, q > 1$ According to the Hölder's inequality, $\dot{\varepsilon}(w)$ satisfies

$$\dot{\varepsilon}(w) \leq \|\nabla_w R(w)\|_q \|w\|_p \|w^{[p-1]}\|_q \|\Delta(w)\|_p - \|\nabla_w R(w)\|_q$$
$$= \|\nabla_w R(w)\|_q - \|\nabla_w R(w)\|_q = \mathbf{0}.$$

The equal sign is taken if and only if $w = \pm \Delta(w)$. Moreover, when $w = \Delta(w)$, $\varepsilon(w)$ has the maximum value which is not stable according to $\dot{\varepsilon}(w) \leq \mathbf{0}$. Thus $w^* = -\Delta(w^*)$ is the asymptotically equilibrium point of PDE (15).

This completes the proof of theorem.

After proving the convergence of $L_p$SGD in PDE form, we can derive the convergence condition of $L_p$SGD about learning rate $\eta$.

**Lemma 1.** When the empirical risk $R$ is a convex function, and its gradient on $w$ satisfies Lipschitz continuity condition in $L_p$ norm space:

$$\|\nabla_w R(w_1) - \nabla_w R(w_2)\|_{p/(p-1)}^{p/(p-1)} < \beta^{p/(p-1)} \|w_1 - w_2\|_p^p, \tag{16}$$

the second-order derivative in any direction is obtained:

$$0 \leq u^T \nabla_w^2 R(w) u < \beta, \tag{17}$$

where $u$ is any vector that satisfies $\|u\|_p = 1$.

*Proof.* Since $R$ is a convex function, the second-order derivative should be non-negative, that means $u^T \nabla_w^2 R(w) u \geq 0$.

Moreover, since the gradient satisfies the Lipschitz continuity condition (16), according to Hölder's inequality, there is

$$\frac{u^T [\nabla_w R(w + ru) - \nabla_w R(w)]}{r} \leq \frac{1}{r} \|u\|_p \|\nabla_w R(w + ru) - \nabla_w R(w)\|_{p/(p-1)} < \beta \|u\|_p^p = \beta.$$

On the other hand, the directional second-order derivative is defined as

$$u^T \nabla_w^2 R(w) u = \lim_{r \to 0} \frac{u^T [\nabla_w R(w + ru) - \nabla_w R(w)]}{r}.$$

Combining the previous two functions, we obtain

$$u^T \nabla_w^2 R(w) u < \beta.$$

This completes the proof of lemma.

**Lemma 2.** When the empirical risk satisfies lemma 1 conditions, the difference between two empirical risks satisfies

$$R(w_2) - R(w_1) < \nabla_w R(w_1)^T (w_2 - w_1) + \frac{\beta}{2} \|w_1 - w_2\|_p^2. \tag{18}$$

*Proof.* Let $d = \|w_2 - w_1\|_p$, $u = \frac{w_2 - w_1}{d}$. We can expand $R(w_2)$ at $w_1$

$$R(w_2) = R(w_1) + \int_0^d \nabla_w R(w_1 + ru)^T u \, dr.$$

Let $g(r) = \nabla_w R(w_1 + ru)^T u$ the equation is rewritten as

$$R(w_2) = R(w_1) + \int_0^d g(r) dr.$$

Since $g'(r) = u^T \nabla_w^2 R(w_1 + ru)^T u$. According to lemma 1, there is $0 \leq g'(r) < \beta$, thus the maximum value of $R(w_2)$ satisfies

$$R(w_2) = R(w_1) + \nabla_w R(w_1)^T (w_2 - w_1) + \int_0^d \int_0^r g'(s) ds dr$$

$$< R(w_1) + \nabla_w R(w_1)^T (w_2 - w_1) + \frac{\beta}{2} \|w_1 - w_2\|_p^2.$$

This completes the proof of lemma.

**Theorem 2.** When the empirical risk $R$ is a convex function, and its gradient on $w$ satisfies the Lipschitz continuity condition (16), under the iteration (13), the convergence condition of sequence $\{w_t\}$ is given by

$$0 < \eta < 2 \frac{\|\nabla_w R(w_t)\|_q - \nabla_w R(w_t)^T w_t \left(w_t^{[p-1]}\right)^T \Delta(w_t)}{\beta \left\| w_t \left(w_t^{[p-1]}\right)^T \Delta(w_t) - \Delta(w_t) \right\|_p^2}, \tag{19}$$

where $\Delta(w_t)$ satisfies the definition (10).

*Proof.* According to lemma 2, $R(w_{t+1}) - R(w_t)$ satisfies

$$R(w_{t+1}) - R(w_t) \le \nabla_w R(w_t)^T (w_{t+1} - w_t) + \frac{\beta}{2} \|w_{t+1} - w_t\|_p^2.$$

When $\eta$ is a small quantity, $w_{t+1} - w_t$ can be expanded at the first order of $\eta$

$$w_{t+1} - w_t \approx \eta \left[ w_t \left(w_t^{[p-1]}\right)^T \Delta(w_t) - \Delta(w_t) \right].$$

Replace $w_{t+1} - w_t$ with this approximate in inequality of $R(w_{t+1})$, then

$$R(w_{t+1}) - R(w_t) \le \eta \nabla_w R(w_t)^T w_t \left(w_t^{[p-1]}\right)^T \Delta(w_t) - \eta \|\nabla_w R(w_t)\|_q +$$
$$\frac{\beta \eta^2}{2} \left\| w_t \left(w_t^{[p-1]}\right)^T \Delta(w_t) - \Delta(w_t) \right\|_p^2.$$

When the sequence $\{w_t\}$ converges, we must make sure that $R(w_{t+1}) < R(w_t)$, thus $\eta$ should satisfies

$$R(w_{t+1}) - R(w_t) \le$$
$$\eta \nabla_w R(w_t)^T w_t \left(w_t^{[p-1]}\right)^T \Delta(w_t) - \eta \|\nabla_w R(w_t)\|_q + \frac{\beta \eta^2}{2} \left\| w_t \left(w_t^{[p-1]}\right)^T \Delta(w_t) - \Delta(w_t) \right\|_p^2 < 0.$$

Then we obtain

$$0 < \eta < 2 \frac{\|\nabla_w R(w_t)\|_q - \nabla_w R(w_t)^T w_t \left(w_t^{[p-1]}\right)^T \Delta(w_t)}{\beta \left\| w_t \left(w_t^{[p-1]}\right)^T \Delta(w_t) - \Delta(w_t) \right\|_p^2}. \tag{20}$$

This completes the proof of theorem.

Theorem 2 implies a phenomenon: when $w$ approaching the optimal solution $w^*$, the upper limit of learning rate $\eta$ should gradually decrease, otherwise $w$ will oscillate around $w^*$. Therefore, we use a substantially decreasing learning rate in $L_p$SGD.

### $L_p$SGD with Momentum

During the batch optimization with $L_p$SGD, the direction of gradient is unstable, that will lead to the oscillation of weight. Since the momentum term can improve the speed of convergence by bringing some eigen components of the system closer to critical damping[23], we introduce the momentum into $L_p$SGD and proposed the $L_p$SGD-m. Moreover, to stabilize the direction of weight vector and increase the control of Hoyer's sparsity, an indirect update of weight is applied in $L_p$SGD-m.

Set the momentum gradient of weight is $\mu_w$, the iteration of weight is given by

$$\mu_{w,j}^{(l,t)} = \gamma \mu_{w,j}^{(l,t-1)} + \nabla_{w_j^{(l)}} R(w_j^{(l,t)}), \tag{21}$$

$$v_j^{(l,t+1)} = \frac{1}{\rho_j^{(l,t+1)}} \left[ (1-\eta) v_j^{(l,t)} - \eta \frac{\mu_{w,j}^{(l,t)}}{\lambda_\mu} \right], \tag{22}$$

and

$$w_j^{(l,t+1)} = \left[ v_j^{(l,t+1)} \right]^{[q-1]}, \tag{23}$$

where $\gamma$ is the momentum, $\boldsymbol{v}$ is the intermediate variable for weight, $\rho_j^{(l,t+1)} = \left\| (1-\eta)\boldsymbol{v}_j^{(l,t)} - \eta \frac{\boldsymbol{\mu}_{w,j}^{(l,t)}}{\lambda_\mu} \right\|_{q^{(l)}}$ is the normalization factor of $\boldsymbol{v}_j^{(l,t+1)}$, and $\lambda_\mu$ satisfies

$$\lambda_\mu = \left\| \boldsymbol{\mu}_{w,j}^{(l,t)} \right\|_q. \tag{24}$$

Set the momentum term of bias is $\mu_b$, the iteration of bias is given by

$$\mu_{b,j}^{(l,t)} = \gamma \mu_{b,j}^{(l,t-1)} + \nabla_{b_j^{(l)}} R(b_j^{(l,t)}), \tag{25}$$

$$b_j^{(l,t+1)} = b_j^{(l,t)} - \eta\, \mu_{b,j}^{(l,t+1)}. \tag{26}$$

The pseudocode of $L_p$SGD-m is summarized on Box 1.

---

**Box 1 | $L_p$-spherical gradient descent with momentum ($L_p$SGD-m) pseudocode is detailed in Algorithm 2**

**Algorithm 2: $L_p$SGD-m pseudocode**

**Input**: Neural network defined in (1), Training dataset $\{x_1,...,x_n\}$, $\{y_1,...,y_n\}$, Learning rate $\eta$, constraint parameters $p^{(l)}$, $l=1,...,L$

**Output**: Parameters of network $\vartheta = \{w_j^{(l)}, b_j^{(l)}, l=1,\cdots,L, j=1,\cdots,m^{(l)}\}$

1: Initialize $\vartheta$, and the normalized gradient defined in (10);
2: **FOR** each training epoch
3:    **FOR** each batch
4:       Sampling *batchSize* number of data form dataset;
5:       **FOR** $l=1, ...,L$
6:          Update the momentum gradient of weight and bias for layer $l$ with (21) and (25)
7:          Update the intermediate variable $\boldsymbol{v}_j^{(l)}$ of layer $l$ with (22)
8:          Update the weights $\boldsymbol{w}_j^{(l)}$ of layer $l$ with (23);
9:          Update the bias $b_j^{(l)}$ of layer $l$ with (26);
10:      **END FOR**
11:    **END FOR**
12:    Reduce the learning rate $\eta$;
13: **END FOR**

---

**Expectation of Hoyer's sparsity under gamma distribution**

Given the training sample $\boldsymbol{x} \in \mathbb{R}^d$, the most activated weight $\boldsymbol{w}$ constrained on $L_p$-sphere maximize the dot product with $\boldsymbol{x}$:

$$\begin{aligned} \boldsymbol{w}^* &= \arg\max \boldsymbol{w} \cdot \boldsymbol{x} \\ s.t.\ &\|\boldsymbol{w}\|_p = 1. \end{aligned} \tag{27}$$

Function (27) is solved as:

$$w^* = \frac{x^{[1/(p-1)]}}{\|x^{[1/(p-1)]}\|_p}, \tag{28}$$

where

$$x^{[t]} = \text{sgn}(x) \circ |x|^t, \tag{29}$$

represent the Hadamard product of $t$-power of absolute value and corresponding sign for each element of $x$, that means the components of $x^{[t]}$ satisfies $x_j^{[t]} = \text{sgn}(x_j)|x_j|^t$. And $\|x\|_p = \sqrt[p]{\sum_{i=1}^d |x_i|^p}$ is the $L_p$-norm of $x$.

Supposing the components of $x$ are independent of each other, and the probability density function (p.d.f.) satisfies following gamma distribution:

$$f(x_k) = \frac{1}{(p-1)\Gamma(\alpha(p-1)/2)} |x_k|^{\alpha-1} \exp(-|x_k|^{2/(p-1)}), k=1,\cdots,d, \alpha \in (1,+\infty), \tag{30}$$

where $\Gamma(\bullet)$ is the gamma function.

According to Hoyer's sparsity defined by (2), we have the expectation of weight sparsity under the p.d.f. $p(x) = \prod f(x_k)$:

$$\begin{aligned}
E_x[H] &= \int p(x) H(w(x)) dx \\
&= \frac{\sqrt{d}}{\sqrt{d}-1} - \frac{1}{(\sqrt{d}-1)(p-1)^d \Gamma^d(\alpha(p-1)/2)} \times \\
&\quad \int_{-1}^{1} \cdots \int_{-1}^{1} \exp\left(-\sum_{k=1}^d x_k^{2/(p-1)}\right) \frac{\|x^{[1/(p-1)]}\|_1}{\|x^{[1/(p-1)]}\|_2} \prod_{k=1}^d |x_k|^{\alpha-1} dx_1 \cdots dx_d.
\end{aligned} \tag{31}$$

Consider the transformation

$$z = |x|^{[1/(p-1)]}. \tag{32}$$

Then the expectation of Hoyer's sparsity defined by (31) is transformed to

$$\begin{aligned}
E_x[H] &= \frac{\sqrt{d}}{\sqrt{d}-1} - \frac{2^d}{(\sqrt{d}-1)\Gamma^d(\alpha(p-1)/2)} \times \\
&\quad \int_{-1}^{1} \cdots \int_{-1}^{1} \exp\left(-\sum_{k=1}^d z_k^2\right) \prod_{k=1}^d z_k^{\alpha(p-1)-1} \frac{\|z\|_1}{\|z\|_2} dz_1 \cdots dz_d.
\end{aligned} \tag{33}$$

Noticing the denominator contains $\|z\|_2$, that causes the difficulty in integration. To eliminate $\|z\|_2$, the Cartesian coordinate should be transferred to hyperspherical coordinate. The Jacobians of the above transformation are given by

$$J(z \to (\theta_1,\cdots,\theta_{d-1},r)) = r^{d-1} \prod_{k=1}^{d-1} \cos^{k-1}\theta_k. \tag{34}$$

Setting $\tau = \alpha(p-1)$. According to the Jacobians transformation (34), the expectation (33) satisfies,

$$E_x[H] = \frac{\sqrt{d}}{\sqrt{d}-1} - \frac{2^d}{(\sqrt{d}-1)\Gamma^d(\tau/2)} \times \int_0^{+\infty} r^{d\tau-1} e^{-r^2} dr \int_0^{\pi/2} \cdots \int_0^{\pi/2} \prod_{k=1}^{d-1} (\sin^{\tau-1}\theta_k \cos^{k\tau-1}\theta_k) \times$$
$$(\sin\theta_{d-1} + \cos\theta_{d-1}(\cdots(\sin\theta_2 + \cos\theta_2(\sin\theta_1 + \cos\theta_1))\cdots))d\theta_1 \cdots d\theta_{d-1} \quad (35)$$
$$= \frac{\sqrt{d}}{\sqrt{d}-1} - \frac{2^{d-1}\Gamma(d\tau/2)}{(\sqrt{d}-1)\Gamma^d(\tau/2)} \times \int_0^{\pi/2} \cdots \int_0^{\pi/2} \prod_{k=1}^{d-1} (\sin^{\tau-1}\theta_k \cos^{k\tau-1}\theta_k) \times$$
$$(\sin\theta_{d-1} + \cos\theta_{d-1}(\cdots(\sin\theta_2 + \cos\theta_2(\sin\theta_1 + \cos\theta_1))\cdots))d\theta_1 \cdots d\theta_{d-1}.$$

For each $\theta_k$ in (35), there are three independent integrals

$$\begin{cases} \Omega_{k,1} = \int_0^{\pi/2} \sin^{\tau-1}\theta_k \cos^{k\tau-1}\theta_k d\theta_k = \frac{1}{k\tau} {}_2F_1\left(1-\frac{\tau}{2}, \frac{k\tau}{2}; \frac{k\tau}{2}+1; 1\right) \\ \Omega_{k,2} = \int_0^{\pi/2} \sin^{\tau}\theta_k \cos^{k\tau-1}\theta_k d\theta_k = \frac{1}{k\tau} {}_2F_1\left(\frac{1-\tau}{2}, \frac{k\tau}{2}; \frac{k\tau}{2}+1; 1\right) \\ \Omega_{k,3} = \int_0^{\pi/2} \sin^{\tau-1}\theta_k \cos^{k\tau}\theta_k d\theta_k = \frac{1}{k\tau+1} {}_2F_1\left(1-\frac{\tau}{2}, \frac{k\tau+1}{2}; \frac{k\tau+3}{2}; 1\right), \end{cases} \quad (36)$$

where ${}_2F_1$ is the hypergeometric functions.

Thus the expectation of Hoyer's sparsity is given by

$$E[H] = \frac{\sqrt{d}}{\sqrt{d}-1} - \frac{2^{d-1}\Gamma(d\tau/2)}{(\sqrt{d}-1)\Gamma^d(\tau/2)} \xi_{d-1}, \quad (37)$$

where $\xi_{d-1}$ satisfies

$$\begin{cases} \xi_1 = \Omega_{1,2} + \Omega_{1,3}, \Pi_1 = \Omega_{1,1} \\ \xi_{k+1} = \xi_k \Omega_{k+1,3} + \Pi_k \Omega_{k+1,2} \\ \Pi_{k+1} = \Pi_k \Omega_{k+1,1}, k = 1, 2, \cdots, d-2. \end{cases} \quad (38)$$

**Implementation of WRN-22 training on CIFAR-10.**

WRN-22s (such as WRN-22-2, WRN-22-4, WRN-22-6 and WRN-22-8) has 21 convolutional layers with 3×3 kernels, 2 convolution layers with 1×1 kernels, and a final fully-connected layer[29]. It consists of three different ResNet blocks with six or eight 3×3 kernels each. After the first convolution layer, there is a unity residual feed forward connection after every two convolution layers, except for an 1×1 residual convolution connection to make output channels compatible between two layers. Each convolution layer is followed by batch normalization[31]. ReLU activation is used after the batch normalization except for the residual connections, where the ReLU activation is computed after summation. Each residual connection has 1×1 convolution and the first convolution of ResNet blocks 2 and 3 downsample the input by using a stride of 2 pixels. The 8×8 output of the last convolution layer is then downsampled to 1×1 resolution using global average pooling[32], which is followed by a single fully-connected layer. For the last fully-connected layer, no batch normalization is performed.

Each WRN-22 is trained for 250 epochs on CIFAR-10 classification dataset which containing 50,000 images for training and 10,000 for test. The loss function is categorical cross entropy function over ten classes of the input image. The model is initialized using He Normal initialization[33]. The basic learning rate is performed in Fig. 9b. In order to accelerate the convergence, WRN-22s are trained with a momentum of 0.9, a batch size of 512 and a drop fraction of 0.2 or relative threshold of 0.1. To overcome the overfitting, images from the train set are firstly randomly cropped to 32x32 patch after padding 4 pixels, then execute a random horizontal flip, and finally cutout on them.

In experiments, the most sparse WRN-22-2, WRN-22-4, WRN-22-6, WRN-22-8 are trained under $p$=1.07 on CIFAR-10. WRN-22-2 contains 1,154,074 parameters, 1,152,352 of them are synaptic weights. WRN-22-4 contains 4,595,290 parameters, 4,591,872 of them are synaptic weights. WRN-22-6 contains 10,324,122 parameters, 10,319,008 of them are synaptic weights. WRN-22-8 contains 18,340,570 parameters, 18,333,760 of them are synaptic weights. To evaluate the sparsity-accuracy trade-off of WRN-22-2, it is trained for several times under different final sparsity with suitable $p$.

**Data availability**

The data used in this paper are public datasets, freely available online from the following link.

UCI machine learning repository: http://archive.ics.uci.edu/ml/index.php.

Where UCI Adult: http://archive.ics.uci.edu/ml/datasets/Adult

UCI Connect-4: http://archive.ics.uci.edu/ml/datasets/Connect-4

UCI DNA:

http://archive.ics.uci.edu/ml/datasets/Molecular+Biology+(Splice-junction+Gene+Sequences)

UCI Mushroom: http://archive.ics.uci.edu/ml/datasets/Mushroom

UCI Letter: http://archive.ics.uci.edu/ml/datasets/Letter+Recognition

UCI Climate: http://archive.ics.uci.edu/ml/datasets/Climate+Model+Simulation+Crashes

UCI Web: http://archive.ics.uci.edu/ml/datasets/Website+Phishing

The MNIST is available at: http://yDNN.lecun.com/exdb/mnist/

The Fashion-MNIST is available at: https://github.com/zalandoresearch/fashion-mnist

CIFAR-10 is available at: http://www.cs.toronto.edu/~kriz/cifar.html

Tiny ImageNet is available at: http://tiny-imagenet.herokuapp.com/

**Code availability**

The demo of $L_p$SS can be found from: https://github.com/WilliamLiPro/LpSS

The code of RigL, SET, SNIP is available from: https://github.com/google-research/rigl

## Acknowledgements

This work was supported by the Natural Science Foundation of China (under grant 61806209), and the Key Laboratory of Shaanxi Province Open Foundation (under grant SKLIIN-20180103). We thank Y. Lan for providing the suggestions for the experiments, and G. Fan and P. Huang for their support with the experimental activities.